\def\year{2018}\relax
\documentclass[letterpaper]{article} 
\usepackage{aaai18}  
\usepackage{times}  
\usepackage{helvet}  
\usepackage{courier}  
\usepackage{url}  
\usepackage{graphicx}  
\usepackage{multirow}
\newcommand{\nop}[1]{}

\usepackage{amsmath, amsfonts, amssymb, xspace, color,listings,enumitem,graphicx}

\newcommand{\our}{\textsc{LM-LSTM-CRF}\xspace}
\newcommand{\ournl}{\textsc{LM-LSTM-CRF\_NL}\xspace}
\newcommand{\ournh}{\textsc{LM-LSTM-CRF\_NH}\xspace}

\newcommand{\saft}{}

\def \myspace {\_}
\def \c {\mathbf{c}}

\def \f {\mathbf{f}}

\def \m {\mathbf{m}}
\def \n {\mathbf{n}}

\def \r {\mathbf{r}}
\def \t {\mathbf{t}}

\def \v {\mathbf{v}}
\def \w {\mathbf{w}}
\def \x {\mathbf{x}}
\def \y {\mathbf{y}}
\def \z {\mathbf{z}}

\def \Y {\mathbf{Y}}
\def \Z {\mathbf{Z}}
\def \fl {\mathbf{f^{L}}}
\def \fw {\mathbf{f^{N}}}
\def \rl {\mathbf{r^{L}}}
\def \rw {\mathbf{r^{N}}}

\begin{document}

\title{Empower Sequence Labeling with Task-Aware Neural Language Model}

\author{
Liyuan Liu$^{\dag}~$
Jingbo Shang$^{\dag}~$
Xiang Ren$^{\sharp}~$ 
Frank F. Xu$^{\ddag}~$
Huan Gui$^{\flat}~$
Jian Peng$^{\dag}~$
Jiawei Han$^{\dag}~$
\\[0.5ex]
{$^{\dag}$ University of Illinois at Urbana-Champaign, \{ll2, shang7, jianpeng, hanj\}@illinois.edu}\\
{$^{\sharp}$ University of Southern California, xiangren@usc.edu} \\
{$^{\ddag}$ Shanghai Jiao Tong University, frankxu@sjtu.edu.cn}\\
{$^{\flat}$ Facebook, huangui@fb.com}
}

\maketitle

\begin{abstract}

Linguistic sequence labeling is a general approach encompassing a variety of problems, such as part-of-speech tagging and named entity recognition.
Recent advances in neural networks (NNs) make it possible to build reliable models without handcrafted features.
However, in many cases, it is hard to obtain sufficient annotations to train these models.
In this study, we develop a neural framework to extract knowledge from raw texts and empower the sequence labeling task.
Besides word-level knowledge contained in pre-trained word embeddings, character-aware neural language models are incorporated to extract character-level knowledge.
Transfer learning techniques are further adopted to mediate different components and guide the language model towards the key knowledge.
Comparing to previous methods, these task-specific knowledge allows us to adopt a more concise model and conduct more efficient training.
Different from most transfer learning methods, the proposed framework does not rely on any additional supervision.
It extracts knowledge from self-contained order information of training sequences.
Extensive experiments on benchmark datasets demonstrate the effectiveness of leveraging character-level knowledge and the efficiency of co-training.
For example, on the CoNLL03 NER task, model training completes in about 6 hours on a single GPU, reaching F$_1$ score of 91.71$\pm$0.10 without using any extra annotations.
\end{abstract} 

\begin{figure*}[ht!]
  \centering
    \includegraphics[width=\textwidth]{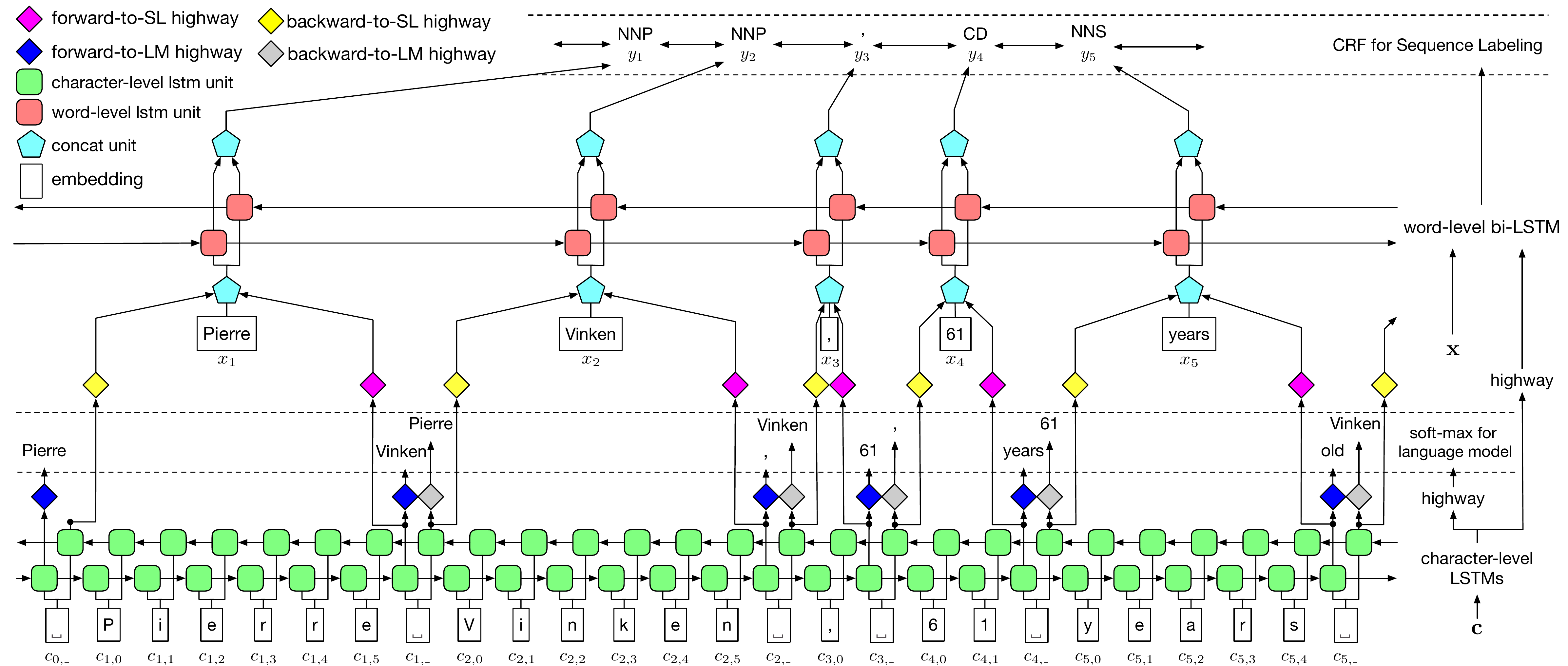}
  \caption{\our Neural Architecture}
  \label{fig:Framework}
\saft
\end{figure*}

\section{Introduction}
\label{sect:intro}

Linguistic sequence labeling is a fundamental framework.
It has been applied to a variety of tasks including part-of-speech (POS) tagging, noun phrase chunking and named entity recognition (NER)~\cite{ma-hovy:2016:P16-1,sha2003shallow}.
These tasks play a vital role in natural language understanding and fulfill lots of downstream applications, such as relation extraction, syntactic parsing, and entity linking~\cite{2017arXiv170700166L,luo2015joint}.

Traditional methods employed machine learning models like Hidden Markov Models (HMMs) and Conditional Random Fields (CRFs), and have achieved relatively high performance.
However, these methods have a heavy reliance on handcrafted features (e.g., whether a word is capitalized) and language-specific resources (e.g., gazetteers).
Therefore, it could be difficult to apply them to new tasks or shift to new domains.
To overcome this drawback, neural networks (NNs) have been proposed to automatically extract features during model learning.
Nevertheless, considering the overwhelming number of parameters in NNs and the relatively small size of most sequence labeling corpus, annotations alone may not be sufficient to train complicated models.
So, guiding the learning process with extra knowledge could be a wise choice.

Accordingly, transfer learning and multi-task learning have been proposed to incorporate such knowledge.
For example, NER can be improved by jointly conducting other related tasks like entity linking or chunking~\cite{luo2015joint,peng2016improving}.
After all, these approaches would require additional supervision on related tasks, which might be hard to get, or not even existent for low-resource languages or special domains.

Alternatively, abundant knowledge can be extracted from raw texts, and enhance a variety of tasks.
Word embedding techniques represent words in a continuous space~\cite{mikolov2013distributed,pennington2014glove} and retain the semantic relations among words.
Consequently, integrating these embeddings 
could be beneficial to many tasks~\cite{2017arXiv170700166L,2016naacl}.
Nonetheless, most embedding methods take a word as a basic unit, thus only obtaining word-level knowledge, while character awareness is also crucial and highly valued in most state-of-the-art NN models.

Only recently, character-level knowledge has been leveraged and empirically verified to be helpful in numerous sequence labeling tasks~\cite{peters2017semi,rei2017semi}.
Directly adopting pre-trained language models, character-level knowledge can be integrated as context embeddings and demonstrate its potential to achieve the state-of-the-art~\cite{peters2017semi}.
However, the knowledge extracted through pre-training is not task-specific, thus containing a large irrelevant portion.
So,
this approach would require a bigger model, external corpus and longer training.
For example, one of its language models was trained on 32 GPUs for more than half a month, which is unrealistic in many situations.

In this paper, we propose an effective sequence labeling framework, \our, 
which leverages both word-level and character-level knowledge in an efficient way.
For character-level knowledge, we incorporate a neural language model with the sequence labeling task
and conduct multi-task learning to guide the language model towards task-specific key knowledge.
Besides the potential of training a better model, this strategy also poses a new challenge.
Based on our experiments, when the tasks are discrepant, language models could be harmful to sequence labeling in a na\"ive co-training setting.
For this reason, we employ highway networks~\cite{srivastava2015highway} to transform the output of character-level layers into different semantic spaces,
thus mediating and unifying these two tasks.
For word-level knowledge, we choose to fine-tune pre-trained word embeddings instead of co-training or pre-training the whole word-level layers,
because the majority of parameters in word-level layers come from the embedding layer and such co-training or pre-training cost lots of time and resources.

We conduct experiments on the CoNLL 2003 NER task, the CoNLL 2000 chunking task, as well as the WSJ portion of the Penn Treebank POS tagging task.
\our achieves a significant improvement over the state-of-the-art.
Also, our co-training strategy allows us to capture more useful knowledge with a smaller network, thus yielding much better efficiency without loss of effectiveness.

\section{\our Framework}
\label{sect:model}

The neural architecture of our proposed framework, \our, is visualized in Fig.~\ref{fig:Framework}.
For a sentence with annotations $\y=(y_1, \dots, y_n )$, its word-level input is marked as $\x=(x_1, x_2, \dots, x_n)$, where $x_i$ is the $i$-th word; its character-level input is recorded as $\c=(c_{0,\myspace}, c_{1, 1}, c_{1, 2}, \dots, c_{1, \myspace}, c_{2, 1}, \dots, c_{n,\myspace})$, where $c_{i, j}$ is the j-th character for word $w_i$ and $c_{i, \myspace}$ is the space character after $w_i$.
These notations are also summarized in Table~\ref{tbl:notation}.

\begin{table}[t]
\center
\begin{tabular}{c|l|c|l}
\hline
$\x$ & word-level input & $x_i$ & $i$-th word \\
\hline
$\c$ & character-level input & $c_{i,j}$ & $j$-th char in $x_i$ \\
\hline
$c_{i, \myspace}$ & space after $x_i$ & $c_{0, \myspace}$ & space before $x_1$\\
\hline
$\y$ & label sequence & $y_i$ & label of $x_i$ \\
\hline
$\f_i$ & \multicolumn{3}{l}{output of forward character-level LSTM at $c_{i, \myspace}$}\\
\hline
$\r_i$ & \multicolumn{3}{l}{output of backward character-level LSTM at $c_{i, \myspace}$}\\
\hline
$\fl_i$ & \multicolumn{3}{l}{output of forward-to-LM highway unit}\\
\hline
$\rl_i$ & \multicolumn{3}{l}{output of backward-to-LM highway unit}\\
\hline
$\fw_i$ & \multicolumn{3}{l}{output of forward-to-SL highway unit}\\
\hline
$\rw_i$ & \multicolumn{3}{l}{output of backward-to-SL highway unit}\\
\hline
$\v_i$ & \multicolumn{3}{l}{input of word-level bi-LSTM at $x_i$}\\
\hline
$\z_i$ & \multicolumn{3}{l}{output of word-level bi-LSTM at $x_i$}\\
\hline
\end{tabular}
\caption{Notation Table.}\label{tbl:notation}
\end{table}

Now, we first discuss the multi-task learning strategy and then introduce the architecture in a bottom-up fashion.

\subsection{Multi-task Learning Strategy}
As shown in Fig.~\ref{fig:Framework}, our language model and sequence labeling model share the same character-level layer, which fits the setting of multi-task learning and transfer learning.
However, different from typical models of this setting, our two tasks are not strongly related.
This discordance makes our problem more challenging.
E.g., although a naive co-training setting, which directly uses the output from character-level layers, could be effective in several scenarios~\cite{yang2017transfer},
for our two tasks, it would hurt the performance.
This phenomenon would be further discussed in the experiment section.

To mediate these two tasks, we transform the output of character-level layers into different semantic spaces 
for different objectives.
This strategy allows character-level layers to focus on general feature extraction and lets the transform layers select task-specific features.
Hence, our language model can provide related knowledge to the sequence labeling, without forcing it to share the whole feature space.

\subsection{Character-level Layer}

Character-level neural language models are trained purely on unannotated sequence data but can capture the underlying style and structure.
For example, it can mimic Shakespeare's writing and generate sentences of similar styles, or even master the grammar of programming languages (e.g., XML, \LaTeX, and C) and generate syntactically correct codes~\cite{rnn-effectiveness-blog}.
Accordingly, we adopted the character-level Long Short Term Memory (LSTM) networks to process character-level input.
Aiming to capture lexical features instead of remembering words' spelling,
we adjust the prediction from the next character to the next word.
As in Fig.~\ref{fig:Framework}, the character-level LSTM would only make predictions for the next word at word boundaries (i.e., space characters or $c_{i, \myspace}$).

Furthermore, we coupled two LSTM units to capture information in both forward and backward directions.
Although it seems similar to the bi-LSTM unit, the outputs of these two units are processed and aligned differently.
Specifically, we record the output of forward LSTM at $c_{i, \myspace}$ as $\f_{i}$, and the output of backward LSTM at $c_{i, \myspace}$ as $\r_{i}$.

\subsection{Highway Layer}

In computer vision, Convolutional Neural Networks (CNN) has been proved to be an effective feature extractor, but its output needs to be further transformed by fully-connected layers to achieve the state-of-the-art.
Bearing this in mind, it becomes natural to stack additional layers upon the flat character-level LSTMs.
More specifically, we employ highway units~\cite{srivastava2015highway}, which allow unimpeded information flowing across several layers.
Typically, highway layers conduct nonlinear transformation as
$
\m = H(\n) = \t \odot g(W_H \n + b_H) + (1-\t) \odot \n
$
, where $\odot$ is element-wise product, $g(\cdot)$ is a nonlinear transformation such as ReLU in our experiments, $\t=\sigma(W_T \n + b_T)$ is called transform gate and $(1-\t)$ is called carry gate.

In our final architecture, there are four highway units, named \texttt{forward-to-LM}, \texttt{forward-to-SL}, \texttt{backward-to-LM}, and \texttt{backward-to-SL}.
The first two transfer $\f_i$ into $\fl_i$ and $\fw_i$, and the last two transfer $\r_i$ into $\rl_i$ and $\rw_i$.
$\fl_i$ and $\rl_i$ are used in the language model, while $\fw_i$ and $\rw_i$ are used in the sequence labeling.

\subsection{Word-level Layer}

Bi-LSTM is adopted as the word-level structure to capture information in both directions.
As shown in Fig.~\ref{fig:Framework}, we concatenate $\fw_i$ and $\rw_{i-1}$ with word embeddings and then feed them into the bi-LSTM.
Note that, in the backward character-level LSTM, $\c_{i-1, \myspace}$ is the space character before word $x_{i}$, therefore, $\fw_i$ would be aligned and concatenated with $\rw_{i-1}$ instead of $\rw_{i}$.
For example, in Fig.~\ref{fig:Framework}, the word embeddings of `Pierre' will be concatenated with the output of the \texttt{forward-to-SL} over `$\ldots$Pierre\myspace' and the output of the \texttt{backward-to-SL} over `$\ldots$erreiP\myspace'.

As to word-level knowledge, we chose to fine-tune pre-trained word embeddings, instead of co-training the whole word-level layer.
This is because most parameters of our word-level model come from word embeddings, and fine-tuning pre-trained word embeddings have been verified to be effective in leveraging word-level knowledge~\cite{ma-hovy:2016:P16-1}.
Besides, current word embedding methods can easily scale to the large corpus; pre-trained word embeddings are available in many languages and domains~\cite{fernandezvecshare}.
However, this strategy cannot be applied to character-level layers, since the embedding layer of character-level layers contains very few parameters.
Based on these considerations, we applied different strategies to leverage word-level knowledge from character-level.

\subsection{CRF for Sequence Labeling}

Label dependencies are crucial for sequence labeling tasks.
For example, in NER task with BIOES annotation, it is not only meaningless but illegal to annotate \texttt{I-PER} after \texttt{B-ORG} (i.e., mixing the person and the organization).
Therefore, jointly decoding a chain of labels can ensure the resulting label sequence to be meaningful.
Conditional random field (CRF) has been included in most state-of-the-art models to capture such information and further avoid generating illegal annotations.
Consequently, we build a CRF layer upon the word-level LSTM.

For training instance $(\x_i, \c_i, \y_i)$, we suppose the output of word-level LSTM is $\Z_i=(\z_{i, 1}, \z_{i, 2}, \dots, \z_{i, n})$.
CRF models describe the probability of generating the whole label sequence with regard to $(\x_i, \c_i)$ or $\Z$.
That is, $p(\hat{\y} | \x_i, \c_i)$ or $p(\hat{\y} | \Z)$, where $\hat{\y} = (\hat{y}_1, \dots, \hat{y}_n)$ is a generic label sequence.
Similar to~\cite{ma-hovy:2016:P16-1}, we define this probability as follows.
\begin{equation}
p(\hat{\y}|\x_i, \c_i) = \frac{ \prod_{j=1}^n \phi(\hat{y}_{j-1}, \hat{y}_j, \z_j) } { \sum_{\y' \in \Y(\Z)} \prod_{j=1}^n \phi(y'_{j-1}, y'_j, \z_j)}
\label{eqn:crf_prob}
\end{equation}
Here, $\Y(\Z)$ is the set of all generic label sequences, $\phi(y_{j-1}, y_j, \z_j) = \exp(W_{y_{j-1}, y_{j}} \z_i + b_{y_{j-1}, y_{j}})$, where $W_{y_{j-1}, y_{j}}$ and $b_{y_{j-1}, y_{j}}$ are the weight and bias parameters corresponding to the label pair $(y_{j-1}, y_{j})$.

For training, we minimize the following negative log-likelihood.
\begin{equation}
\mathcal{J}_{CRF} = -\sum_i \log p(\y_i | \Z_i)
\label{eqn:crf}
\end{equation}
And for testing or decoding, we want to find the optimal sequence $\y^*$ that maximizes the likelihood.
\begin{equation}
\y^* = \arg \underset{\y \in \Y(\Z)}{{\max}} \; p(\y | \Z)
\label{eqn:decode}
\end{equation}
Although the denominator of Eq.~\ref{eqn:crf_prob} is complicated, we can calculate Eqs.~\ref{eqn:crf} and \ref{eqn:decode} efficiently by the Viterbi algorithm.

\subsection{Neural Language Model}

The language model is a family of models describing the generation of sequences.
In a neural language model, the generation probability of the sequence $\x = (x_1, ..., x_n)$ in the forward direction (i.e., from left to right) is defined as
$$
p_f(x_1, ..., x_n) = \prod_{i=1}^N p_f(x_i| x_1, \dots, x_{i-1})
$$
where $p_f(x_i| x_1, \dots, x_{i-1})$ is computed by NN.

In this paper, our neural language model makes predictions for words but takes the character sequence as input. Specifically, we would calculate $p_f(x_i| c_{0, \myspace}, \dots, c_{i-1, 1}, \dots, c_{i-1, \myspace})$ instead of $p_f(x_i | x_1, \dots, x_{i-1})$. This probability is assumed as
$$
p_f(x_i| c_{0, \myspace}, \dots, c_{i-1, \myspace}) = \frac{\exp(\w_{x_i}^T \fw_{i-1})}{\sum_{\hat{x}_j} \exp(\w_{\hat{x}_j}^T \fw_{i-1})}
$$
where $\w_{x_i}$ is the weight vector for predicting word $x_i$.
In order to extract knowledge in both directions, we also adopted a reversed-order language model, 
which calculates the generation probability from right to left as
\begin{align*}
&p_r(x_1, ..., x_n) = \prod_{i=1}^N p_r(x_i| c_{i+1, \myspace}, \dots, c_{n, \myspace}) \\
\mbox{where }& p_r(x_i| c_{i+1, \myspace}, \dots, c_{n, \myspace}) = \frac{\exp(\w_{x_i}^T \rw_{i})}{\sum_{\hat{x}_j} \exp(\w_{\hat{x}_j}^T \rw_{i})}
\end{align*}
The following negative log likelihood is applied as the objective function of our language model.
\begin{equation}
\mathcal{J}_{LM} = -\sum_{i} \log p_f(\x_i) -\sum_{i} \log p_r(\x_i)
\label{eqn:lm}
\end{equation}

\subsection{Joint Model Learning}

By combining Eqs.~\ref{eqn:crf} and \ref{eqn:lm}, we can write the joint objective function as
\begin{equation}
\mathcal{J} = -\sum_{i} \Big( p(\y_i | \Z_i) + \lambda \big( \log p_f(\x_i) + \log p_r(\x_i) \big) \Big)
\label{eqn:obj}
\end{equation}
where $\lambda$ is a weight parameter. In our experiments, $\lambda$ is always set to $1$ without any tuning.

In order to train the neural network efficiently, stochastic optimization has been adopted. And at each iteration, we sample a batch of training instances and perform an update according to the summand function of Eq.~\ref{eqn:obj}: $
p(\y_i | \Z_i) + \lambda \big( \log p_f(\x_i) + \log p_r(\x_i) \big)
$

\begin{table}[t]
\center
\begin{tabular}{c||c|c|c}
\hline
\multirow{2}{*}{Dataset}& \multicolumn{3}{|c}{\# of Sentences} \\
\cline{2-4}
&  Train & Dev & Test\\
\hline
\hline
\textbf{CoNLL03 NER} & 14,987 & 3,466 & 3,684 \\
\hline
\textbf{CoNLL00 chunking} & 7,936 & 1,000 & 2,012 \\
\hline
\textbf{WSJ} & 38,219 & 5,527 & 5,426 \\
\hline
\end{tabular}
\caption{Dataset summary.}\label{tbl:dataset}
\saft
\end{table}

\section{Experiments}
\label{sect:exp}

Here, we evaluate \our on three benchmark datasets: the CoNLL 2003 NER dataset~\cite{tjong2003introduction}, the CoNLL 2000 chunking dataset~\cite{tjong2000introduction}, and the Wall Street Journal portion of Penn Treebank dataset (WSJ)~\cite{marcus1993building}.
\begin{itemize}[noitemsep,nolistsep]
\item \textbf{CoNLL03 NER} contains annotations for four entity types: \texttt{PER}, \texttt{LOC}, \texttt{ORG}, and \texttt{MISC}. It has been separated into training, development and test sets.
\item \textbf{CoNLL00 chunking} defines eleven syntactic chunk types (e.g., \texttt{NP}, \texttt{VP}) in addition to \texttt{Other}. It only includes training and test sets. Following previous works~\cite{peters2017semi}, we sampled 1000 sentences from training set as a held-out development set.
\item \textbf{WSJ} contains 25 sections and categorizes each word into 45 POS tags. We adopt the standard split and use sections 0-18 as training data, sections 19-21 as development data, and sections 22-24 as test data~\cite{manning2011part}.
\end{itemize}

The corpus statistics are summarized in Table~\ref{tbl:dataset}. 
We report the accuracy for the WSJ dataset.
And in the first two datasets, we adopt the official evaluation metric (micro-averaged F$_1$), and use the BIOES scheme
~\cite{ratinov2009design}.
Also, in all three datasets, rare words (i.e., frequency less than 5) are replaced by a special token ($<$UNK$>$).

\subsection{Network Training}

\begin{table}
\center
\scalebox{0.95}
{
\begin{tabular}{c|c|c|c|c}
\hline
Layer & Parameter & POS & NER & chunking \\
\hline
\hline
character-level & \multirow{2}{*}{dimension} & \multicolumn{3}{|c}{\multirow{2}{*}{30}} \\
embedding & & \multicolumn{3}{}{}\\
\hline
character-level & depth & \multicolumn{3}{|c}{1} \\
\cline{2-5}
LSTM & state size & \multicolumn{3}{|c}{300} \\
\hline
Highway & depth & \multicolumn{3}{|c}{1} \\
\hline
word-level & \multirow{2}{*}{dimension} & \multicolumn{3}{|c}{\multirow{2}{*}{100}} \\
embedding & & \multicolumn{3}{}{}\\
\hline
word-level & depth & \multicolumn{3}{|c}{1}\\
\cline{2-5}
bi-LSTM & state size & \multicolumn{3}{|c}{300} \\
\hline
Optimization & $\eta_0$ & 0.015 & \multicolumn{2}{|c}{0.01}\\
\hline
\end{tabular}
}
\caption{Hyper-parameters of \our.}\label{tbl:hyper}
\saft
\end{table}

For a fair comparison, we didn't spend much time on tuning parameters but borrow the initialization, optimization method, and all related hyper-parameter values (except the state size of LSTM) from the previous work~\cite{ma-hovy:2016:P16-1}.
For the hidden state size of LSTM, we expand it from $200$ to $300$, because introducing additional knowledge allows us to train a larger network. 
We will further discuss this change later. 
Table~\ref{tbl:hyper} summarizes some important hyper-parameters. 
Since the CoNLL00 is similar to the CoNLL03 NER dataset, we conduct experiments with the same parameters on both tasks.

\noindent\textbf{Initialization.} We use GloVe $100$-dimension pre-trained word embeddings released by Stanford\footnote{\url{http://nlp.stanford.edu/projects/glove/}} and randomly initialize the other parameters~\cite{glorot2010understanding,jozefowicz2015empirical}. 

\noindent\textbf{Optimization.} We employ mini-batch stochastic gradient descent with momentum. 
The batch size, the momentum and the learning rate are set to $10$, $0.9$ and $\eta_t = \frac{\eta_0}{1 + \rho t}$, where $\eta_0$ is the initial learning rate and $\rho = 0.05$ is the decay ratio. 
Dropout is applied in our model, and its ratio is fixed to $0.5$. 
To increase stability, we use gradient clipping of $5.0$.

\noindent\textbf{Network Structure.} The hyper-parameters of character-level LSTM are set to the same value of word-level bi-LSTM.
We fix the depth of highway layers as $1$ to avoid an over-complicated model.

Note that some baseline methods (e.g., \cite{Chiu2016NamedER,peters2017semi}) incorporate the development set as a part of training. However, because we are using early stopping based on the evaluation on the development set, our model is trained purely on the training set. 

\subsubsection{Compared Methods}
We consider three classes of baseline sequence labeling methods in our experiments.
\begin{itemize}[noitemsep,nolistsep]
    \item \textbf{Sequence Labeling Only.}
    Without any additional supervision or extra resources, LSTM-CRF~\cite{2016naacl} and LSTM-CNN-CRF~\cite{ma-hovy:2016:P16-1} are the current state-of-art methods.
    We also list some top reported performance on each dataset~\cite{Collobert2011NaturalLP,luo2015joint,Chiu2016NamedER,yang2017transfer,peters2017semi,manning2011part,sogaard2016deep,sun2014structure}.

    \item \textbf{Joint Model with Other Supervised Tasks.}
    There are several attempts~\cite{luo2015joint,yang2017transfer} to enhance sequence labeling tasks by introducing additional annotations from other related tasks (e.g., enhance NER with entity linking labels). 

    \item \textbf{Joint Model with Language Model}:
    Language models have been employed by some recent works to extract knowledge from raw text and thus enhancing sequence labeling task. 
    TagLM~\cite{peters2017semi} leverages pre-trained language models and shows the effectiveness with the large external corpus, 
    but the large model scale and long training time make it hard to re-run this model.
    Another work \cite{rei2017semi} also incorporates the sequence labeling task with the language model.

\end{itemize}
For comparison, we tune the parameters of three most related baselines~\cite{ma-hovy:2016:P16-1,2016naacl,rei2017semi}\footnote{Implementations: \url{https://github.com/xuezhemax/lasagnenlp} (Ma et al. 2016), \url{https://github.com/glample/tagger} (Lample et al. 2016) and \url{https://github.com/marekrei/sequence-labeler} (Rei 2017)}.
    , and report the statics of the best working parameter setting.
Besides, we index these models by number, and summarize the results in Tables~\ref{tbl:ner},~\ref{tbl:pos}~and~\ref{tbl:chunking}.

\begin{table}[t!]
\center
\scalebox{0.7}
{
\begin{tabular}{c||l|l|l}
\hline
\multirow{2}{*}{Extra Resource} & \multirow{2}{*}{Index \& Model} & \multicolumn{2}{|c}{F$_1$ score}\\
\cline{3-4}
 & & Type & Value ($\pm$std)\\
\hline
\hline
\multirow{2}{*}{gazetteers} & 0) Collobert et al. 2011$^\dagger$ & reported& 89.59 \\
\cline{2-4}
& 1) Chiu et al. 2016$^\dagger$ & reported & 91.62$\pm$0.33 \\
\hline
AIDA dataset & 2) Luo et al. 2015 & reported& 91.20 \\
\hline
CoNLL 2000 / & \multirow{2}{*}{3) Yang et al. 2017$^\dagger$} & \multirow{2}{*}{reported} & \multirow{2}{*}{91.26} \\
PTB-POS dataset & & & \\
\hline
1B Word dataset \& & \multirow{2}{*}{4) Peters et al. 2017$^\dagger$$^\ddagger$ } & \multirow{2}{*}{reported} & \multirow{2}{*}{91.93$\pm$0.19} \\
\texttt{4096-8192-1024} & & & \\
\hline
1B Word dataset & 5) Peters et al. 2017$^\dagger$$^\ddagger$ & reported & 91.62$\pm$0.23 \\
\hline
\multirow{15}{*}{None} & 6) Collobert et al. 2011$^\dagger$ & reported& 88.67 \\
\cline{2-4}
& 7) Luo et al. 2015 & reported& 89.90 \\
\cline{2-4}
& 8) Chiu et al. 2016$^\dagger$ & reported & 90.91$\pm$0.20 \\
\cline{2-4}
& 9) Yang et al. 2017$^\dagger$ & reported & 91.20\\
\cline{2-4}
& 10) Peters et al. 2017$^\dagger$ & reported & 90.87$\pm$0.13\\
\cline{2-4}
& 11) Peters et al. 2017$^\dagger$$^\ddagger$ & reported & 90.79$\pm$0.15\\
\cline{2-4}
& \multirow{3}{*}{ 12) Rei 2017 $^\dagger$$^\ddagger$ } & mean & 87.38$\pm$0.36\\
\cline{3-4}
& & max & 87.94\\
\cline{3-4}
& & reported & 86.26\\
\cline{2-4}
& \multirow{3}{*}{13) Lample et al. 2016$^\dagger$} & mean & 90.76$\pm$0.08\\
\cline{3-4}
& & max & 91.14\\
\cline{3-4}
& & reported & 90.94\\
\cline{2-4}
& \multirow{3}{*}{14) Ma et al. 2016$^\dagger$} & mean & 91.37$\pm$0.17\\
\cline{3-4}
& & max & 91.67\\
\cline{3-4}
& & reported & 91.21\\
\cline{2-4}
& \multirow{2}{*}{15) \our$^\dagger$$^\ddagger$} & mean & 91.71$\pm$0.10 \\
\cline{3-4}
& & max & 91.85\\
\hline
\end{tabular}
}
\caption{F$_1$ score on the CoNLL03 NER dataset.
We mark models adopting pre-trained word embedding as $\dagger$, and record models which leverage language models as $\ddagger$.
}\label{tbl:ner}
\end{table}

\subsection{Performance Comparison}
In this section, we focus on the comparisons between \our and previous state-of-the-arts, including both effectiveness and efficiency.
As demonstrated in Tables~\ref{tbl:ner},~\ref{tbl:pos}~and~\ref{tbl:chunking}, \our significantly outperforms all baselines without additional resources. 
Moreover, even for those baselines with extra resources, \our beats most of them and is only slightly worse than TagLM (index 4)~\cite{peters2017semi}.

TagLM (index 4) is equipped with both extra corpoa (about $4000X$ larger than the CoNLL03 NER dataset) and a tremendous pre-trained forward language model (\texttt{4096-8192-1024}\footnote{\texttt{4096-8192-1024} is composed of character-level CNN with $4096$ filters, $2$ layers of stacked LSTMs with $8192$ hidden units each and a $1024$-dimension projection unit.})~\cite{jozefowicz2016exploring}.
Due to the expensive resources and time required by \texttt{4096-8192-1024}, even the authors of TagLM failed to train a backward language model of the same size, instead, chose a much smaller one (\texttt{LSTM-2048-512}\footnote{\texttt{LSTM-2048-512} is composed of a single-layer LSTM with $2048$ hidden units and a $512$-dimension projection unit.}).
It is worth noting that, when either extra corpus or \texttt{4096-8192-1024} is absent, \our shows significant improvements over TagLM (index 5, 10 and 11). 

Also, LSTM-CNN-CRF outperforms LSTM-CRF in our experiments, which is different from~\cite{reimers2017reporting}.
During our experiments, we discover that, when trained on CPU, LSTM-CNN-CRF only reaches 90.83 F$_1$ score on the NER dataset, but gets 91.37 F$_1$ score when trained on GPU.
We conjecture that this performance gap is due to the difference of runtime environments.
Therefore, we conduct all of our experiments on GPU. 
Additionally, we can observe that, although co-trained with language model, results of index 12 fails to outperform LSTM-CNN-CRF or LSTM-CRF.
The reason of this phenomenon could be complicated and beyond the scope of this paper.
However, it verified the effectiveness of our method, and demonstrated the contribution of outperforming these baselines.

    \subsubsection{NER}
    First of all, we have to point out that the results of index 1, 4, 8, 10 and 11 are not directly comparable with others since their final models are trained on both training and development set, while others are trained purely on the training set. 
    As mentioned before, \our outperforms all baselines except TagLM (index 4).
    For a thorough comparison, we also compare to its variants, TagLM (index 5), TagLM (index 10) and TagLM (index 11).
    Both index 10 and 11 are trained on the CoNLL03 dataset alone, while index 11 utilizes language model and index 10 doesn't.
    Comparing F$_1$ scores of these two settings, we can find that TagLM (index 11) even performs worse than TagLM (index 10) , which reveals that directly applying co-training might hurt the sequence labeling performance.
    We will also discuss this challenge later in the Highway Layers \& Co-training section.

    Besides, changing the forward language model from \texttt{4096-8192-1024} to \texttt{LSTM-2048-512}, TagLM (index 5) gets a lower F$_1$ score of 91.62$\pm$0.23.
    Comparing this score to ours (91.71$\pm$0.10), one can verify that pre-trained language model usually extracts a large portion of unrelated knowledge.
    Relieving such redundancy by guiding the language model with task-specific information, our model is able to conduct both effective and efficient learning.

\begin{table}
\center
\scalebox{0.9}
{
\begin{tabular}{l||l|l}
\hline
\multirow{2}{*}{Ind \& Model} & \multicolumn{2}{|c}{Accuracy}\\
\cline{2-3}
& Type & Value ($\pm$std)\\
\hline
\hline
0) Collobert et al. 2011$^\dagger$ & reported& 97.29\\
\hline
16) Manning 2011 & reported & 97.28 \\
\hline
17) \nocite{sogaard2011semisupervised} S{\o}gaard 2011& reported& 97.50\\
\hline
18) \nocite{sun2014structure} Sun 2014 & reported& 97.36 \\
\hline
\multirow{3}{*}{12) Rei 2017$^\dagger$$^\ddagger$} & mean & 96.97$\pm$0.22\\
\cline{2-3}
& max & 97.14\\
\cline{2-3}
& reported & 97.43\\
\hline
\multirow{2}{*}{13) Lample et al. 2016$^\dagger$} & mean$\pm$std & 97.35$\pm$0.09\\
\cline{2-3}
& maximum & 97.51\\
\hline
\multirow{3}{*}{14) Ma et al. 2016$^\dagger$} &mean$\pm$std & 97.42$\pm$0.04\\
\cline{2-3}
& maximum & 97.46\\
\cline{2-3}
& reported & 97.55\\
\hline
\multirow{2}{*}{15) \our$^\dagger$$^\ddagger$} & mean$\pm$std & 97.53$\pm$0.03\\
\cline{2-3}
& maximum & 97.59\\
\hline
\end{tabular}
}
\caption{Accuracy on the WSJ dataset.
We mark models adopting pre-trained word embedding as $\dagger$, and record models which leverage language models as $\ddagger$.
}\label{tbl:pos}
\end{table}

\begin{table}
\center
\scalebox{0.65}
{
\begin{tabular}{c||c|c||c|c||c|c}
\hline
\multirow{2}{*}{Model} & \multicolumn{2}{|c||}{CoNLL03 NER} & \multicolumn{2}{|c||}{WSJ POS} & \multicolumn{2}{|c}{CoNLL00 Chunking} \\
\cline{2-7}
& h & F$_1$Score & h & Accuracy & h & F$_1$Score\\
\hline
\hline
LSTM-CRF & 46 & 90.76 & 37 & 97.35 & 26 & 94.37 \\
\hline
LSTM-CNN-CRF & 7 & 91.22 & 21 & 97.42 & 6 & 95.80 \\
\hline
\our & 6 & 91.71 & 16 & 97.53 & 5 & 95.96 \\
\hline
LSTM-CRF$^\star$ & 4 & 91.19 & 8 & 97.44 & 2 & 95.82 \\
\hline
LSTM-CNN-CRF$^\star$ & 3 & 90.98 & 7 & 96.98 & 2 & 95.51 \\
\end{tabular}
}
\caption{Training statistics of TagLM (index 4 and 5) and \our on the CoNLL03 NER dataset.}\label{tbl:eff2}
\end{table}

    \subsubsection{POS Tagging}
    Similar to the NER task, \our outperforms all baselines on the WSJ portion of the PTB POS tagging task. 
    Although the improvements over LSTM-CRF and CNN-LSTM-CRF are less obvious than those on the CoNLL03 NER dataset, considering the fact that the POS tagging task is believed to be easier than the NER task and current methods have achieved relatively high performance, this improvement could still be viewed as significant.
    Moreover, it is worth noting that for both NER and POS tagging tasks, \our achieves not only higher F$_1$ scores, but also with smaller variances, which further verifies the superiority of our framework.

    \subsubsection{Chunking}
    In the chunking task, \our also achieves relatively high F$_1$ scores, but with slightly higher variances.
    Considering the fact that this corpus is much smaller than the other two (only about $1/5$ of WSJ or $1/2$ of CoNLL03 NER), we can expect more variance due to the lack of training data.
    Still, \our outperforms all baselines without extra resources, and most of the baselines trained with extra resources.

\begin{table}
\center
\scalebox{0.7}
{
\begin{tabular}{c||l|l|l}
\hline
\multirow{2}{*}{Extra Resource} & \multirow{2}{*}{Ind \& Model} & \multicolumn{2}{|c}{F$_1$ score}\\
\cline{3-4}
 & & Type & Value ($\pm$std)\\
\hline
\hline
\multirow{2}{*}{PTB-POS} & \nocite{hashimoto2016joint}19) Hashimoto et al. 2016$^\dagger$ & reported& 95.77\\
\cline{2-4}
& \nocite{sogaard2016deep}20) S{\o}gaard et al. 2016$^\dagger$& reported & 95.56 \\
\hline
CoNLL 2000 / & \multirow{2}{*}{3)Yang et al. 2017$^\dagger$} & \multirow{2}{*}{reported} & \multirow{2}{*}{95.41} \\
PTB-POS dataset & & & \\
\hline
1B Word dataset &4) Peters et al. 2017$^\dagger$$^\ddagger$ & reported & 96.37$\pm$0.05 \\
\hline
\hline
\multirow{10}{*}{None} & \nocite{hashimoto2016joint}21) Hashimoto et al. 2016$^\dagger$ & reported& 95.02 \\
\cline{2-4}
& \nocite{sogaard2016deep}22) S{\o}gaard et al. 2016$^\dagger$& reported & 95.28 \\
\cline{2-4}
&9) Yang et al. 2017$^\dagger$ & reported & 94.66\\
\cline{2-4}
& \multirow{3}{*}{12) Rei 2017$^\dagger$$^\ddagger$} & mean & 94.24$\pm$0.11\\
\cline{3-4}
& & max & 94.33\\
\cline{3-4}
& & reported & 93.88\\
\cline{2-4}
& \multirow{2}{*}{13) Lample et al. 2016$^\dagger$} & mean & 94.37$\pm$0.07\\
\cline{3-4}
& & maximum & 94.49\\
\cline{2-4}
& \multirow{2}{*}{14) Ma et al. 2016$^\dagger$} & mean & 95.80$\pm$0.13\\
\cline{3-4}
& & maximum & 95.93\\
\cline{2-4}
& \multirow{2}{*}{15) \our$^\dagger$$^\ddagger$} & mean & 95.96$\pm$0.08\\
\cline{3-4}
& & maximum & 96.13\\
\hline
\end{tabular}
}
\caption{F$_1$ score on the CoNLL00 chunking dataset. 
We mark models adopting pre-trained word embedding as $\dagger$, and record models which leverage language models as $\ddagger$.
}\label{tbl:chunking}
\end{table}

\begin{table}
\center
\scalebox{0.65}
{
\begin{tabular}{l|l|l|l l}
\hline
Ind \& Model & F$_1$score & Module &\multicolumn{2}{|c}{Time $\cdot$ Device}\\
\hline
\hline
15) \our & 91.71 & total & 6 &h$\cdot$GTX 1080\\
\hline
\multirow{2}{*}{5) Peters et al. 2017} & \multirow{2}{*}{91.62} & \texttt{LSTM-2048-512} & 320 & h$\cdot$Telsa K40\\
\cline{3-5}
& & \texttt{LSTM-2048-512}& 320 &h$\cdot$Telsa K40\\
\hline
\multirow{2}{*}{4) Peters et al. 2017} & \multirow{2}{*}{91.93} & \texttt{4096-8192-1024} & 14112 &h$\cdot$Telsa K40\\
\cline{3-5}
& & \texttt{LSTM-2048-512}& 320 &h$\cdot$Telsa K40\\
\hline
\end{tabular}
}
\caption{Training time and performance of LSTM-CRF, LSTM-CNN-CRF and \our on three datasets. Our re-implementations are marked with $^\star$}\label{tbl:eff1}
\end{table}

    \subsubsection{Efficiency}

    We implement LM-LSTM-CRF\footnote{\url{https://github.com/LiyuanLucasLiu/LM-LSTM-CRF}} based on the PyTorch library\footnote{\url{http://pytorch.org/}}. 
    Models has been trained on one GeForce GTX 1080 GPU, with training time recorded in Table~\ref{tbl:eff1}.

    In terms of efficiency, the language model component in \our only introduces a small number of parameters in two highway units and a soft-max layer, which may not have a very large impact on the efficiency.
    To control variables like infrastructures, we further re-implemented both baselines, and report their performance together with original implementations.
    From the results, these re-implementations achieve better efficiency comparing to the original ones, but yield relative worse performance.
    Also, \our achieves the best performance, and takes twice the training time of the most efficient model, LSTM-CNN-CRF$^\star$.
    Empirically, considering the difference among the implementations of these models, we think these methods have roughly the same efficiency.
    
    Besides, we list the required time and resources for pre-training model index 4 and 5 on the NER task in Table~\ref{tbl:eff2}~\cite{jozefowicz2016exploring}.
    Comparing to these language models pre-trained on external corpus, our model has no such reliance on extensive corpus, and can achieve similar performance with much more concise model and efficient training.
    It verifies that our \our model can effectively leverage the language model to extract task-specific knowledge to empower sequence labeling.

\subsection{Analysis}

    To analyze the performance of \our, we conduct additional experiments on the CoNLL03 NER dataset. 

    \subsubsection{Hidden State Size}

\begin{table}
\center
\scalebox{0.65}
{
\begin{tabular}{c|l|c|c|c}
\hline
Model & State Size & F$_1$score$\pm$std &  Recall$\pm$std &Precision$\pm$std \\
\hline
\hline
\multirow{3}{*}{\our} & 300 &  91.71$\pm$0.10 &92.14$\pm$0.12 & 91.30$\pm$0.13 \\
\cline{2-5}
& 200 & 91.63$\pm$0.23 &92.07$\pm$0.22 & 91.19$\pm$0.30 \\
\cline{2-5}
& 100 & 91.13$\pm$0.32 &91.60$\pm$0.37 & 90.67$\pm$0.32 \\

\hline
\hline
\multirow{3}{*}{LSTM-CRF} & 300 & 90.76$\pm$0.08 &90.82$\pm$0.08 & 90.69$\pm$0.08 \\
\cline{2-5}
& 200 & 90.41$\pm$0.07 &90.63$\pm$0.07 & 90.20$\pm$0.07 \\
\cline{2-5}
& 100 & 90.74$\pm$0.22 &91.08$\pm$0.50 & 90.42$\pm$0.17 \\

\hline
\hline
\multirow{3}{*}{LSTM-CNN-CRF} & 300  & 91.22$\pm$0.19 & 91.70$\pm$0.16 & 90.74$\pm$0.27\\
\cline{2-5}
& 200  & 91.37$\pm$0.17 & 91.08$\pm$0.53 & 90.58$\pm$0.11\\
\cline{2-5}
& 100 & 91.18$\pm$0.10 & 91.56$\pm$0.16 & 90.81$\pm$0.15\\
\hline
\end{tabular}
}
\caption{Effect of hidden state size of LSTM}\label{tbl:size}
\end{table}

    To explore the effect of model size, we train our model with different hidden state sizes.
    For comparison, we also apply the same hidden state sizes to LSTM-CRF and LSTM-CNN-CRF.
    From Table~\ref{tbl:size}, one can easily observe that the F$_1$ score of \our keeps increasing when the hidden state size grows, while LSTM-CNN-CRF has a peak at state size $200$ and LSTM-CRF has a drop at state size $200$. 
    This phenomenon further verified our intuition of employing the language model to extract knowledge and prevent overfitting.

    \subsubsection{Highway Layers \& Co-training}

    To elucidate the effect of language model\footnote{the perplexities of the forward language model on CoNLL03 NER's training / development / test sets are 52.87 / 55.03 / 50.22.} and highway units, we compare \our with its two variants, \ournl and \ournh.
    The first keeps highway units, but optimizes $\mathcal{J}_{CRF}$ alone; the second jointly optimizes $\mathcal{J}_{CRF}$ and $\mathcal{J}_{LM}$, but without highway units.
    As shown in Table~\ref{tbl:hw}, \ournh yields worse performance than \ournl.
    This observation accords with previous comparison between TagLM (index 10) and TagLM (index 11) on the CoNLL03 NER dataset.
    We conjecture that it is because the NER task and the language model is not strongly related to each other.
    In summary, our proposed co-training strategy is effective and introducing the highway layers is necessary. 

\section{Related Work}
\label{sect:rel}

There exist two threads of related work regarding the topics in this paper, which are sequence labeling and how to improve it with additional information. 

\noindent\textbf{Sequence Labeling.}
As one of the fundamental tasks in NLP, linguistic sequence labeling, including POS tagging, chunking, and NER, has been studied for years.
Handcrafted features were widely used in traditional methods like CRFs, HMMs, and maximum entropy classifiers~\cite{Lafferty2001ConditionalRF,McCallum2003EarlyRF,Florian2003NamedER,Chieu2002NamedER}, but also make it hard to apply them to new tasks or domains.
Recently, getting rid of handcrafted features, there are attempts to build end-to-end systems for sequence labeling tasks, such as BiLSTM-CNN~\cite{Chiu2016NamedER}, LSTM-CRF~\cite{2016naacl}, and the current state-of-the-art method in NER and POS tagging tasks, LSTM-CNN-CRF~\cite{ma-hovy:2016:P16-1}.
These models all incorporate character-level structure, and report meaningful improvement over pure word-level model.
Also, CRF layer has also been demonstrated to be effective in capturing the dependency among labels.
Our model is based on the success of LSTM-CRF model and is further modified to better capture the char-level information in a language model manner.

\begin{table}
\center
\scalebox{0.65}
{
\begin{tabular}{c|l|c|c|c}
\hline
State Size & Model & F$_1$score$\pm$std &  Recall$\pm$std &Precision$\pm$std \\
\hline
\hline
\multirow{3}{*}{300} & \our & 91.71$\pm$0.10 &92.14$\pm$0.12 & 91.30$\pm$0.13 \\
\cline{2-5}
& \ournl & 91.43$\pm$0.09 & 91.85$\pm$0.18 & 91.01$\pm$0.19 \\
\cline{2-5}
& \ournh & 91.16$\pm$0.22 & 91.67$\pm$0.28 & 90.66$\pm$0.23 \\
\hline
\hline

\multirow{3}{*}{200} & \our & 91.63$\pm$0.23 &92.07$\pm$0.22 & 91.19$\pm$0.30 \\
\cline{2-5}
& \ournl & 91.44$\pm$0.10 & 91.95$\pm$0.16 & 90.94$\pm$0.16 \\
\cline{2-5}
& \ournh & 91.34$\pm$0.28 & 91.79$\pm$0.18 & 90.89$\pm$0.30 \\
\hline
\hline
\multirow{3}{*}{100} & \our & 91.13$\pm$0.32 &91.60$\pm$0.37 & 90.67$\pm$0.32 \\
\cline{2-5}
& \ournl & 91.17$\pm$0.11 & 91.72$\pm$0.14 & 90.61$\pm$0.21 \\
\cline{2-5}
& \ournh & 91.01$\pm$0.19 & 91.50$\pm$0.21 & 90.53$\pm$0.30 \\
\hline

\end{tabular}
}
\caption{Effect of language model and highway}\label{tbl:hw}
\end{table}

\noindent\textbf{Leveraging Additional Information.}
Integrating word-level and character-level knowledge has been proved to be helpful to sequence labeling tasks.
For example, word embeddings~\cite{mikolov2013distributed,pennington2014glove} can be utilized by co-training or pre-training strategies~\cite{2017arXiv170700166L,2016naacl}.
However, none of these models utilizes the character-level knowledge.
Although directly adopting character-level pre-trained language models could be helpful~\cite{peters2017semi}.
Such pre-trained knowledge is not task-specific and requires a larger neural network, external corpus, and longer training. 
Our model leverages both word-level and character-level knowledge through a co-training strategy, which leads to a concise, effective, and efficient neural network.
Besides, unlike other multi-task learning methods, our model has no reliance on any extra annotation~\cite{peters2017semi} or any knowledge base~\cite{shang2017automated}. 
Instead, it extracts knowledge from the self-contained order information.

\section{Conclusion}
\label{sect:con}

In this paper, we proposed a sequence labeling framework, \our, which effectively leverages the language model to extract character-level knowledge from the self-contained order information.
Highway layers are incorporated to overcome the discordance issue of the naive co-training
Benefited from the effectively captured such task-specific knowledge, we can build a much more concise model, thus yielding much better efficiency without loss of effectiveness (achieved the state-of-the-art on three benchmark datasets) . 
In the future, we plan to further extract and incorporate knowledge from other ``unsupervised'' learning principles and empower more sequence labeling tasks.

\section*{Acknowledgments}
\label{sect:ack}

We thank Junliang Guo, Cheng Cheng and all reviewers for comments on earlier drafts that led to substantial improvements in the final version.
Research was sponsored in part by the U.S. Army Research Lab. under Cooperative Agreement No. W911NF-09-2-0053 (NSCTA), National Science Foundation IIS 16-18481, IIS 17-04532, and IIS-17-41317, grant 1U54GM114838 awarded by NIGMS through funds provided by the trans-NIH Big Data to Knowledge (BD2K) initiative (www.bd2k.nih.gov), and Google PhD Fellowship. 
The views and conclusions contained in this document are those of the author(s) and should not be interpreted as representing the official policies of the U.S. Army Research Laboratory or the U.S. Government. The U.S. Government is authorized to reproduce and distribute reprints for Government purposes notwithstanding any copyright notation hereon.

\bibliographystyle{aaai}
\bibliography{cited}
\end{document}